\documentclass[runningheads]{llncs}

\usepackage{dialogue}
\usepackage{tabularx}
\usepackage{float}
\usepackage{amsmath}
\usepackage{subcaption} 
\usepackage[T1]{fontenc}
\usepackage{graphicx}
\usepackage{multirow}
\usepackage{enumitem}

\begin{document}
\title{LLM-Generated Feedback Supports Learning If Learners Choose to Use It}
\author{Danielle R. Thomas\orcidID{0000-0001-8196-3252}
\and Conrad Borchers\orcidID{0000-0003-3437-8979}
\and Shambhavi Bhushan\orcidID{0009-0004-3695-2334}
\and Erin Gatz\orcidID{0000-0002-6880-5740}
\and Shivang Gupta\orcidID{0000-0002-5713-3782}
\and Kenneth R. Koedinger\orcidID{0000-0002-5850-4768}
}

\authorrunning{D. R. Thomas et al.}
\institute{
Carnegie Mellon University\\
\email{\{drthomas,cborchers,shivangg,koedinger\}@cmu.edu}\\
\email{\{shambhab,egatz\}@andrew.cmu.edu}
}
\maketitle 

\begin{abstract}

 Large language models (LLMs) are increasingly used to generate feedback, yet their impact on learning remains underexplored, especially compared to existing feedback methods. This study investigates how on-demand LLM-generated explanatory feedback influences learning in seven scenario-based tutor training lessons. Analyzing over 2,600 lesson completions from 885 tutor learners, we compare posttest performance among learners across three groups: learners who received feedback generated by \texttt{gpt-3.5-turbo}, those who declined it, and those without access. All groups received non-LLM corrective feedback. To address potential selection bias—where higher-performing learners may be more inclined to use LLM feedback—we applied propensity scoring. Learners with a higher predicted likelihood of engaging with LLM feedback scored significantly higher at posttest than those with lower propensity. After adjusting for this effect, two out of seven lessons showed statistically significant learning benefits from LLM feedback with standardized effect sizes of 0.28 and 0.33. These moderate effects suggest that the effectiveness of LLM feedback depends on the learners' tendency to seek support. Importantly, LLM feedback did not significantly increase completion time, and learners overwhelmingly rated it as helpful. These findings highlight LLM feedback's potential as a low-cost and scalable way to improve learning on open-ended tasks, particularly in existing systems already providing feedback without LLMs. This work contributes open datasets, LLM prompts, and rubrics to support reproducibility.

\end{abstract}

\section{Introduction}
Feedback is one of the most powerful methods to enhance learning \cite{hattie2007power,wisniewski2020power}. Although extensive research has examined the effects of feedback created by humans in educational settings \cite{hattie2007power}, comparatively little attention has been paid to the role of feedback generated by generative AI, in the form of large language models (LLMs), particularly within scenario-based online learning.

The effectiveness of feedback depends on several factors, including timing, type, and task complexity \cite{mertens2022effects}. Among the most impactful forms of feedback is informational feedback—guidance that not only evaluates a learner’s response but also provides actionable suggestions for improvement \cite{hattie2007power}. Research indicates that feedback is more effective when it provides correct, as opposed to incorrect, responses and is immediate, as opposed to delayed \cite{hattie2007power}. Elaborated feedback, such as providing an explanation, yields larger effects on learning than feedback on the correctness of an answer or revealing the answer itself \cite{mertens2022effects}. This study builds on these insights by examining the impact of immediate, \textit{explanatory} feedback within online learning with the feedback generated by LLMs.

Recent research on the use of generative AI, such as LLMs, for feedback generation has demonstrated promising results in educational contexts. For example, the use of ChatGPT-generated hints has been shown to produce learning gains comparable to human tutor-authored help in improving math skills \cite{pardos2024chatgpt}. Similarly, a randomized controlled trial of high-school English language learners found that LLM-generated feedback led to improved essay revision scores and increased motivation compared to students who did not receive such feedback \cite{meyer2024using}. Research on LLM feedback is in its infancy, with previous work showing that LLM-generated content adequately mirrors human-generated content \cite{denny2023can,escalante2023ai}, but little is known about learners' perceptions and the actual learning effects. 

This study focuses on LLM-generated help in the form of explanatory feedback with \textit{tutors as learners} participating in online training. Tutor learners receive explanatory feedback on their open responses. Uniquely, it involves developing tutoring skills instead of focusing on student math learning or writing, which adds to the novelty of this work. To assess the impact and efficiency of LLM-generated feedback, or simply \textit{LLM feedback}, we investigate the following research questions, focusing on both learning outcomes and learners’ perceptions:
\begin{itemize}[leftmargin=0pt]
\item[] \textbf{RQ1:} What differences exist in tutor learning, as evidenced by posttest performance, among tutor learners receiving LLM-generated explanatory feedback within instruction compared to learners not receiving it?  
\item[] \textbf{RQ2:} How does learners’ accuracy and efficiency compare among those who receive LLM-generated feedback compared to those who do not receive LLM-generated feedback? 
\item[] \textbf{RQ3:} Among learners who received the LLM-generated feedback and rated it, what were their perceptions of the feedback, i.e., \textit{helpful}, \textit{not helpful}, or \textit{incorrect}? 

\end{itemize}

\section{Related Work}

\subsection{LLM-Generated Feedback on Learning}

Automated feedback using techniques, such as natural language processing (NLP), has demonstrated positive \cite{cavalcanti2021automatic,demszky2021can} and negative \cite{gurung2023identification,kinder2025effects} learning effects. More recently, general-purpose LLMs such as Claude \cite{caruccio2024claude} and GPT models \cite{achiam2023gpt} have shown promise in generating high-quality feedback for learners. ChatGPT has been found to effectively generate detailed, coherent, and process-focused feedback to university students, aligning well with human instructors in providing feedback that may support students' learning skills \cite{dai2023can}. A randomized controlled trial found that ChatGPT-generated feedback improved the quality of pre-service teachers’ writing compared to human expert-written feedback \cite{kinder2025effects}. Pre-service teachers in the LLM-generated feedback condition spent more time processing the feedback and wrote longer texts, indicating deeper engagement with both the feedback and the task. Moreover, they rated the feedback as more useful and interesting than static feedback. However, it remains unclear which specific components of the LLM-generated feedback contributed to these perceptions. %

Closer to this work, involving \textit{ tutors as learners}, binary coding systems using \texttt{GPT-4o} to generate templated feedback to tutors have shown performance matching humans \cite{lin2024can1}. Sequence labeling has been used to identify elements of correct and incorrect responses to assist with feedback generation \cite{lin2024can2}. Although previous work has focused mainly on the technical implementation of LLM feedback generation, much less attention has been paid to ensuring a robust pedagogical structure that improves learning \cite{stamper2024enhancing}. The present work aims to address this gap by investigating the effectiveness of LLM-generated explanatory feedback on tutor learning and tutor perceptions of the feedback.

\subsection {Scenario-based Lesson Design}

Scenario-based learning allows learners to develop skills in real-world contexts through guided practice \cite{bardach2021power}. This situational learning offers low-risk environments for teachers and tutors to build expertise \cite{thomas2024tutors,thomas2024learning,thompson2019teacher}. Previous studies have shown approximately 20\% learning gain from pretest to posttest in key tutoring skills, such as giving effective praise and responding to student errors \cite{thomas2023tutor}. We employ a modified \textit{predict-observe-explain} model, shown in Fig. 1, which is a cyclical instructional approach that structures learning through experience \cite{gibbs1988learning}. First, tutor learners are presented with a scenario (e.g., a student making a math error) asking them to \textit{predict} how to best respond within 1) an open-response question and 2) a multiple-choice question (MCQ). Learners assigned the treatment condition have the option to receive LLM-generated feedback on their open response (shown in red). Next, they \textit{explain} their reasoning through another set of questions: 3) an open-ended question and 4) a MCQ. The learners then 5) \textit{observe} research-based recommendations and receive delayed corrective feedback (shown in blue) before they 6) \textit{explain} the reasoning behind what they observed. Finally, learners complete a posttest (receiving no feedback), in which they are presented an analogous tutoring scenario and complete the same sequence of predicting the best responses and explaining their rationale (7-10). The posttest contains four possible points (two MCQs and two open-response questions), which are used to compare the performance of learners under treatment and control conditions.

\subsection{Providing Explanatory AI-Generated Feedback}

Past work on automated tutor feedback used NLP, including sequence labeling, to identify key elements in responses and generate templated feedback \cite{lin2024can2,xu2025improving}. These approaches show moderate success but struggle with tutoring nuances, as many skills defy simple correct/incorrect categorization, and key elements often misalign with correctness. This present study takes a different approach, incorporating a predefined schema to provide LLM-generated rephrasing, enabling automated assessment and feedback. To enhance the accuracy of its evaluations, the model is primed using few-shot prompting. These examples enable the model, namely \texttt{GPT-3.5-turbo}, to recognize elements of high-quality tutor responses while identifying common misconceptions.

\begin{figure}[ht]
    \centering
    \includegraphics[width=0.9\textwidth]{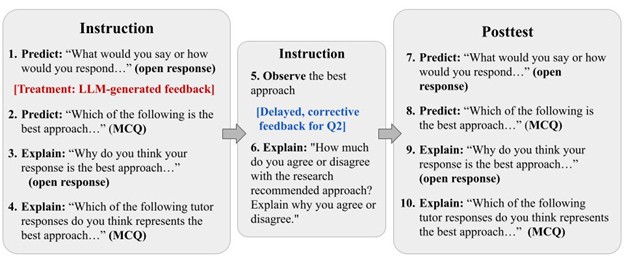}
    \caption{Scenario-based lesson design displaying the option of tutors in the treatment to receive LLM-generated feedback (shown in red) and all participants receiving delayed, corrective feedback (shown in blue) during instruction.}
    \label{fig:lesson design}
\end{figure}

When a tutor learner submits a response, such as the one shown in Fig. 2, the system classifies it as either \textit{correct} or \textit{incorrect} based on its alignment with the predefined schema (e.g., research-backed tutoring strategies). \textit{Correct} responses, which demonstrate effective use of instructional strategies, receive reinforcement feedback that acknowledges their alignment with best practices. In contrast, \textit{incorrect} responses trigger the generation of targeted feedback that not only identifies weaknesses, but also provides a rephrased version generated by the LLM that adheres better to the research-aligned tutoring skill. The model is explicitly instructed to make minimal modifications to incorrect responses, retaining the original structure and intent while improving clarity, specificity, or alignment with instructional goals. Fig. 2 illustrates a tutor inputting an incorrect open response to predict the best approach (i.e., \textit{“Aaron, I see you made a mistake in the first step”}). In line with research-supported effective feedback practices, system-generated feedback suggests how the response could be improved (that is, \textit{ ``...by avoiding explicit mention of the student's mistake''}) and provides an alternative correct response (that is, \textit{...``Let's take a closer look together at the first step''}). Tutors can then attempt the problem again, up to three times, and can also provide feedback on the usefulness of the feedback. 
  
\begin{figure}[ht]
    \centering
    \includegraphics[width=0.75\textwidth]{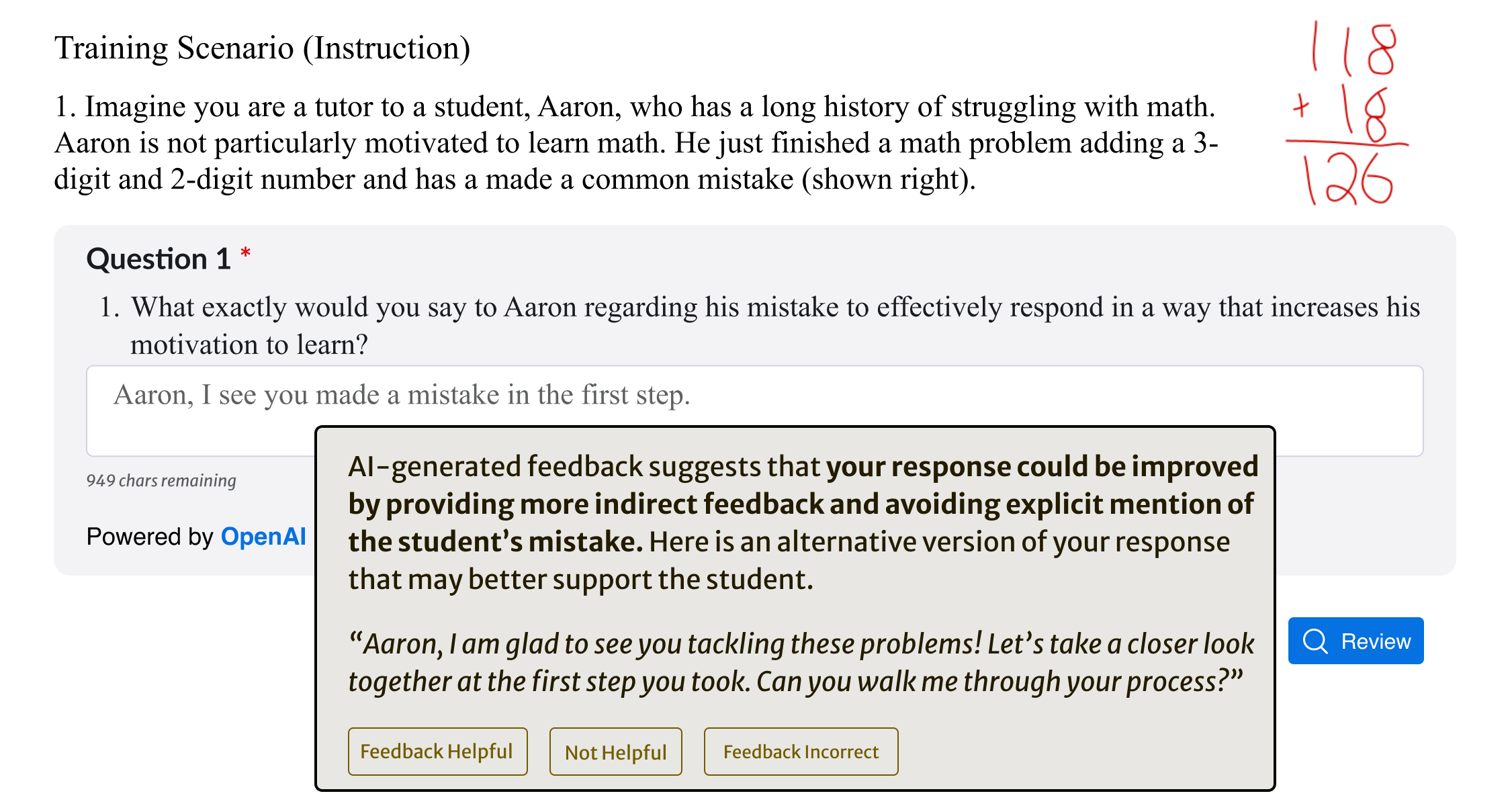}
    \caption{Example of a human tutor receiving LLM-generated, explanatory feedback on their open response (from the \textit{Reacting to Errors} lesson).}
    \label{fig:feedback sample}
\end{figure}

\section{Methods}

\subsection{Participants \& Setting}

A total of 885 college-student tutors accounting for 2,648 lesson completions, completed at least one of the seven lessons titled \textit{Giving Effective Praise}, \textit{Reacting to Errors}, \textit{Determining What Students Know}, \textit{Helping Students Manage Inequity}, \textit{Using Motivational Strategies}, \textit{Supporting a Growth Mindset}, and \textit{Responding to Negative Self-Talk}. The typical participant completed an average of M = 3.42 lessons (SD = 2.13). Delivered through an online tutoring platform, these lessons align with key tutoring competencies \cite{chhabra2022evaluation}. The tutor lesson data was recorded in DataShop, a widely used repository of educational data in learning analytics research \cite{koedinger2010data}. Participant privacy was maintained in compliance with IRB protocols. Lesson log data are accessible through the GitHub.\footnote{https://github.com/conradborchers/ai-feedback-exp/}

\subsection{Research Design \& Analysis Plan}
Learners were randomly assigned to one of two conditions for each lesson: the \textbf{Control} condition, where no LLM-generated feedback was available, or the \textbf{Intent-to-Treat (ITT)} condition, where they had the option to receive LLM-generated explanatory feedback on how to best respond during instruction (Q1 in Fig. 1). Use of LLM feedback was on-demand. The learners assigned the ITT condition who chose to receive LLM-generated feedback on Q1 formed a third condition, the \textbf{Treatment-on-the-Treated (TOT)} condition.  

The on-demand use of LLM feedback within the ITT group introduces a self-selection effect. As a result, the study incorporates both TOT and ITT analyses. For some lessons, all learners were assigned to the ITT condition for a period, leading to an unequal distribution of participants between the ITT and the Control condition. Learners were assigned conditions at lesson level, meaning the same learner could experience various conditions across lessons. This within-subject design reduces variability and enhances statistical power by allowing learners to serve as their own controls.

In addressing RQ1 within the TOT, we examined the learning differences, measured at posttest performance, between the learners who received LLM feedback (\textit{TOT}) to those in the \textit{Control} and \textit{ITT} conditions who opt not to use the LLM feedback. We used a mixed linear model using scaled posttest scores as the outcome variable. The primary predictor was whether the learner received LLM feedback. To account for hierarchical dependencies, random intercepts for learners and lessons were included. This approach controlled for individual and contextual variability while isolating the LLM feedback effect.  
 
In addressing RQ1 within the ITT, we investigated the impact of LLM feedback availability on tutor learning. We conducted an \textit{Intent-to-Treat (ITT)} analysis, comparing tutors assigned to the LLM feedback condition with those in the \textit{Control}. Posttest performance, expressed as standard deviations, served as the primary outcome variable. A linear mixed model was used to estimate the effect of LLM feedback assignment, regardless of whether the tutors opted to receive feedback. Random intercepts for both learners and lessons account for hierarchical dependencies and individual variability. Given that some tutors in the ITT condition did not receive LLM feedback--either by choice or due to system failures--we applied \textit{principal stratification} to control for potential selection bias \cite{sales2019role}. We engineered engagement-related features to predict the likelihood that a learner will request LLM feedback. An elasticNet regression model with 10-fold cross-validation was used to infer feedback propensity \cite{zou2005regularization}. The resulting predictive model was applied to the Control to create a propensity-matched comparison. This approach allowed us to estimate the effect of LLM feedback more accurately by comparing tutors who were likely to request LLM feedback (had it been available) to those who actually received it. We tested the interaction between feedback propensity and condition to assess whether LLM feedback was particularly beneficial for high-propensity learners.

\subsection{Assessing Tutor Learner Posttest Performance}

Tutor posttest performance on each lesson was calculated by summing the binary score (i.e., 0/1) for Q7-Q10 (see Fig. 1), yielding a maximum score of four points. Q8 and Q10 were multiple choice. Q7 and Q9 were open responses and were evaluated using \texttt{GPT-4o}. Each lesson included two scenarios: one presented during instruction and one at posttest. To control for difficulty, the scenarios were randomly assigned and counterbalanced between the participants.\footnote{We note that, in experimental learning research, posttest performance is often used as a proxy for learning. By comparing posttest scores across randomly assigned conditions—which is expected, on average, to even out prior knowledge differences—researchers can infer differences in the amount of learning that occurred during the intervention. This approach assumes that higher post-test performance reflects greater knowledge acquisition attributable to the experimental manipulation.}

\textbf{Human Annotation, Interrater Reliability, and Item-level Reliability.} To evaluate learner’s open responses at posttest (Q7 and Q9 in Fig. 1), we implemented a few-shot learning approach using learner-generated responses in conjunction with human-scored examples to help generate accurate assessments. The prompts were designed to evaluate \textit{predict} (Q7) and \textit{explain} (Q9) open-response types. Prompt development was an iterative process, refined through multiple feedback cycles. Strategies included context framing (e.g., \textit{“\ldots You are a tutor evaluator…”}), incorporating human-coded response examples, and using chain-of-thought prompting to encourage reasoning. To ensure consistency, the model’s temperature was set to 0, and responses were capped at 300 tokens to maintain conciseness. Inter-rater reliability (IRR) was established in prior open-source work for most lessons. For lessons where IRR was not previously established, 50 tutor responses were randomly chosen, and two experienced human coders scored the responses. The IRR for all lessons ranged from 0.64-0.88 for \textit{predict} and 0.65-0.91 for \textit{explain} open responses. Annotation rubrics and lesson-specific reliability are provided in the GitHub. Three MCQ items were found to be unreliable and removed: \textit{Determining What Students Know} (explain); \textit{Reacting to Errors} (explain); and \textit{Responding to Negative Self-Talk} (predict). The results were robust against the exclusion of these items.\footnote{The results were similar and robust with or without the exclusion of three unreliable items, which improved reliability by over 0.05 in Cronbach’s alpha.}

\section{Results}

The descriptive statistics are shown in Table \ref{tab:itt_tot_summary} showing the average posttest score by condition, the number of learners, and the average completion time. 
The average posttest scores where the TOT condition was the highest among conditions are shown in bold. In five of the seven lessons, the TOT condition had the highest average posttest score compared to the other conditions.

\begin{table}[]
\centering
\caption{Lesson Descriptive Statistics. Average posttest scores where the TOT condition was the highest among conditions are shown in bold.}
\label{tab:itt_tot_summary}
\renewcommand{\arraystretch}{1.6} %
\resizebox{1\textwidth}{!}{ %
\begin{tabular}{l|ll|l}
\hline
\multirow{2}{*}[-2mm]{\centering \textbf{Lesson}}                   & \multicolumn{2}{c|}{\textbf{\begin{tabular}[c]{@{}c@{}}Feedback Offered: Intent-to-Treat (ITT)\\ Was LLM Feedback Used?\end{tabular}}} & \multicolumn{1}{c}{\textbf{\begin{tabular}[c]{@{}c@{}}No LLM Feedback\\ Assigned (Control)\end{tabular}}} \\ \cline{2-3}
                                          & \multicolumn{1}{c|}{\textbf{Yes (TOT)}}               & \multicolumn{1}{c|}{\textbf{No Feedback Used}} &  \\ \hline
Effective Praise         & \multicolumn{1}{l|}{\textbf{89.57\%} (76), 7.9 min}    & 85.95\% (120), 8.38 min                & 87.92\% (151), 11.42 min                                               \\ \hline
Manage Inequity & \multicolumn{1}{l|}{82.89\% (38), 8.58 min}   & 83.88\% (76), 6.87 min                 & 81.46\% (120), 10.05 min                                               \\ \hline
What Students Know   & \multicolumn{1}{l|}{\textbf{82.67\%} (25), 4.31 min}   & 80.56\% (102), 5.23 min                & 78.52\% (97), 8.55 min                                                 \\ \hline
Reacting to Errors              & \multicolumn{1}{l|}{64.24\% (188), 10.44 min} & 56.48\% (386), 8.9 min                 & 73.4\% (94), 14.16 min                                                 \\ \hline
Negative Self-Talk & \multicolumn{1}{l|}{\textbf{75.76\%} (33), 5.54 min}   & 72.69\% (72), 7.05 min                 & 65.01\% (121), 9.67 min                                                \\ \hline
Growth Mindset      & \multicolumn{1}{l|}{\textbf{81.88\%} (63), 6.28 min}   & 75.21\% (118), 7.81 min                & 77.18\% (84), 7.62 min                                                 \\ \hline
Motivational Strategies             & \multicolumn{1}{l|}{\textbf{79.94\%} (145), 10.66 min} & 78.35\% (229), 11.08 min               & 78.33\% (75), 7.58 min                                                 \\ \hline
\end{tabular}
}
\end{table}

\subsection{RQ1: Impact of LLM-Generated Feedback on Learning}   

\subsubsection{Treatment-on-the-Treated (TOT) Condition vs No Treatment Assigned or Used.} 
To address RQ1, we examined differences in tutor learning, as evidenced by posttest performance, among learners \textit{actually} receiving LLM-generated feedback compared to those who did not (combined Control and ITT, who opted not to receive LLM feedback.) We used a linear mixed model based on scaled posttest scores, expressed in standard deviations. The primary predictor of interest was whether students received LLM-generated feedback. Random intercepts were included for both learners and lessons to account for the hierarchical structure and variability at these levels. This approach allowed us to model individual and contextual factors while isolating the effect of LLM feedback on performance.

The results indicate that receiving LLM feedback was associated with significantly higher posttest performance, with an estimated effect of 0.10 SD (95\% CI [0.01, 0.19], \textit{p}=.023). The intraclass correlation coefficient (ICC) showed that 26.9\% of the variability in posttest performance was attributable to individual learner differences and 8.8\% to lesson-level differences. To further investigate the impact of LLM-generated feedback at the lesson level, we fitted models for individual lessons to estimate effect sizes. Effect size estimates ranged from 0.02 to 0.27 SD, with the largest effect observed in the \textit{Responding to Negative Self-Talk} (Estimate=0.27) and the smallest in the \textit{Helping Students Manage Inequity} (Estimate=0.02). However, sample sizes within individual lessons were insufficient to reliably test these effects with adequate statistical power.

\textbf{Adjusting for Selection Bias via Principal Stratification and Propensity Modeling.} Approximately one-third (33.2\%) of lesson completions in the assigned LLM-generated feedback condition did not receive feedback, either because the learners did not request it or due to occasional system failures, raising potential bias in estimating effects. To address this, we used principal stratification to create a more fair control group that resembles the tutors likely to request LLM feedback with similar tutors in the control group. Following \cite{sales2019role}, we engineered features to match tutors with a high proclivity to request LLM feedback to similar tutors in the control, specifically leveraging metrics related to general engagement, response behaviors, and session characteristics during tutor training relevant to both conditions. These features included the total attempts of each tutor in the questions and the unique problems attempted. Time-based metrics such as the total duration of training, the average duration, and the range of attempt durations were also considered.

Additionally, we examined input length of features (e.g., min, max, and average input length) of open-ended responses and session-based navigation behaviors (e.g., average time between attempts). We then used these characteristics to predict the tendency of a tutor to request LLM feedback under the intervention condition, represented by the total number of times they had requested LLM-generated feedback. We then performed a 10-fold cross-validation using an ElasticNet regression model to infer the number of times a tutor requested LLM feedback in the intervention condition. The ElasticNet model regularizes parameters for variable selection using weighted absolute value (L1) and quadratic (L2) penalties to shrink estimates toward 0. Both the weighting of both penalties and the strength of the penalty can be independently tuned \cite{zou2005regularization}. During cross-validation, we tuned hyperparameters alpha (the weight of L1 and L2 penalties) and lambda (the penalty strength). Following \cite{sales2022geepers}, we also averaged cross-validation ROC-AUC through a median split of labels and prediction in correctly predicting this outcome across folds, achieving satisfactory accuracy across folds (M = 0.75), which is fair for this method \cite{sales2022geepers}. We then applied this prediction model to the control condition, generating a covariate that adjusts for selection effects by estimating the number of LLM feedback requests learners in the control condition would likely have made. Effectively, this process then adjusts learners' posttest scores who requested and received generative LLM feedback to those our model \textit{predicted to have used} LLM feedback in the control condition, counteracting potential selection effects. More details on the range of features and code from the statistical analyses can be accessed in the GitHub. Table 2 shows model results after including the predicted number of LLM feedback requests as a covariate.

\begin{table}[ht]
\centering
\caption{Model coefficients of the LLM feedback intervention effect after adjusting for learners’ predicted propensity to request feedback. All coefficients and posttest scores are standardized to a mean of 0 and standard deviation of 1.}
\label{tab:coefficients_summary}
\begin{tabular}{lccc}
\hline
\textbf{} & \textbf{Estimates} & \textbf{95\% CI} & \textbf{p-value} \\
\hline
Intercept         & 0.05               &  -0.18 - 0.28     & .666           \\
Intent-to-Treat (ITT)
              & -0.08             & -0.17 - 0.01    & .102            \\
Control Propensity to Request LLM Feedback           & 0.06              & 0.00 - 0.12     & \textbf{.040}            \\
ITT Interaction with Propensity
       & 0.04               & -0.04 - 0.12      & .307           \\
\hline
\end{tabular}
\end{table}

The model (Table 2) revealed a significant selection effect, where the Control condition's predicted propensity to request LLM feedback was positively associated with posttest scores ($\beta$ = 0.06, 95\% CI [0.00, 0.12], $p = .040$). A positive but non-significant interaction between the ITT condition and propensity to request feedback ($\beta$ = 0.04, 95\% CI [-0.04, 0.12], $p = .307)$ indicated that learners engaging in LLM-generated feedback benefited from it, though not significantly so. There were no significant condition differences, in line with our random assignment ($\beta$ = -0.08, 95\% CI [-0.17, 0.01], $p = .102)$, though the negative trend could signal imprecision in the stratification method used to generate the propensity covariate. Looking at lessons individually, we gathered some evidence that LLM feedback contributed to performance, even after accounting for selection effects. The propensity-adjusted treatment effects (corresponding to the ITT interaction effect with propensity in Table \ref{tab:coefficients_summary}), estimated through separate regressions for each lesson, indicate that two lessons—\textit{Giving Effective Praise} ($\beta = 0.33$) and \textit{Supporting a Growth Mindset} ($\beta = 0.28$)—demonstrate small to moderate, statistically significant effects ($p < .05$). No other treatment effects were statistically significant, ranging from -0.15 $SD$ (\textit{Helping Students Manage Inequity}) to 0.20 $SD$ (\textit{Using Motivational Strategies}).

Notably, simply being assigned the possibility of requesting LLM-generated feedback (ITT) did not significantly improve posttest performance, as the main effect of the ITT condition was not significant (see previous paragraph). This suggests that the potential to request LLM-generated feedback alone was insufficient to enhance performance outcomes, highlighting the importance of actual engagement with LLM-generated feedback to realize its benefits. Fig. 3 illustrates the posttest scores by condition, adjusting for the learner's propensity to request LLM feedback to account for treatment selection effects. Among learners who were offered LLM feedback (blue), those who were more likely to use the LLM feedback were predicted to score higher at posttest. Learners who were not offered LLM feedback (red) did not show much difference in posttest score regardless of their predicted likelihood of using the feedback. This aligns with a positive treatment effect after adjusting for self-selection bias.

\begin{figure}[ht]
    \centering
    \includegraphics[width=0.6\textwidth]{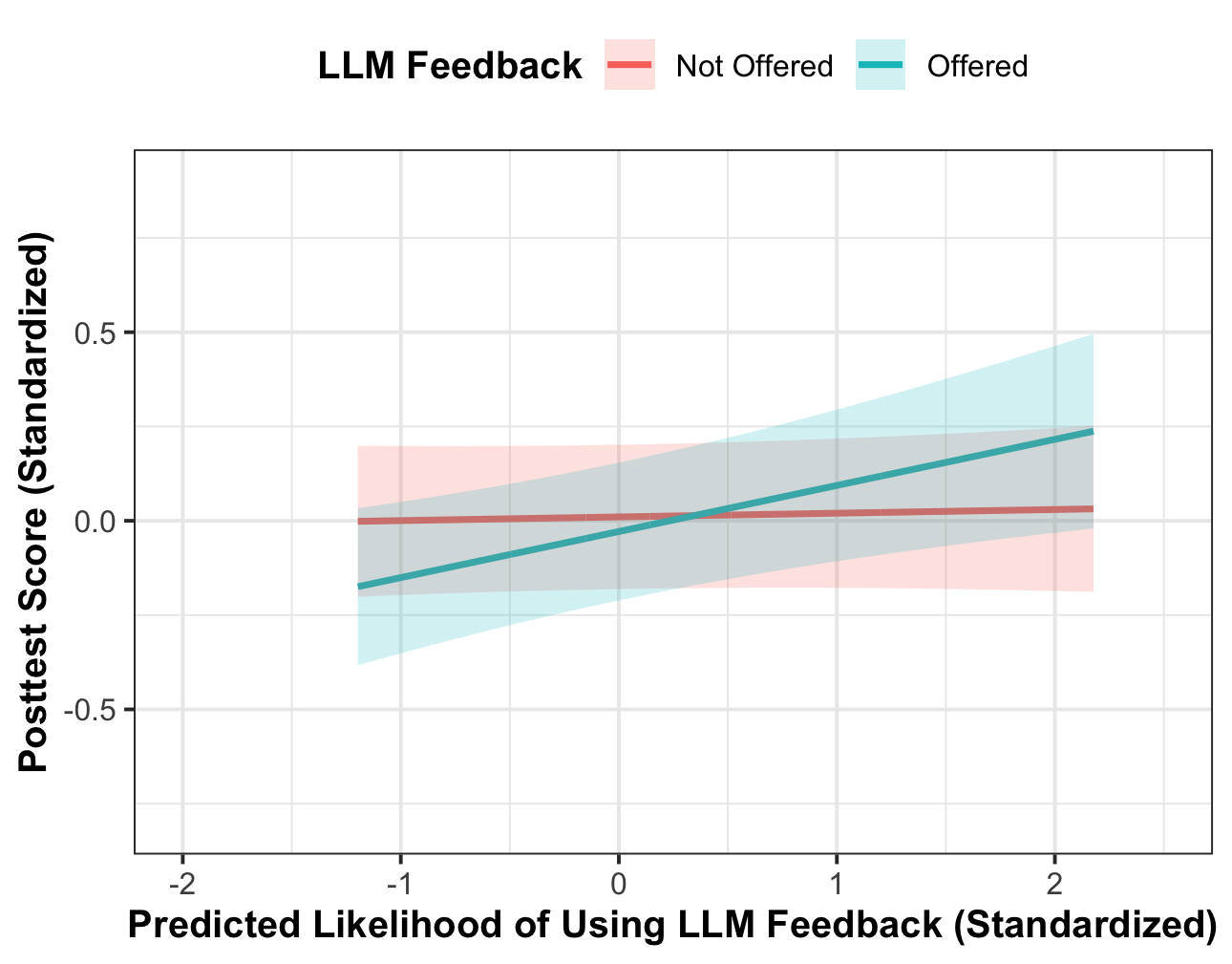}
    \caption{Posttest scores by condition, adjusting for learner propensity to request LLM feedback to adjust for treatment selection effects, including 95\% confidence intervals estimated through model coefficient standard errors.}
    \label{fig:propensity}
\end{figure}

\subsection{RQ2: Influence of LLM Feedback on Lesson Efficiency}
To address RQ2, we investigated the influence of LLM-generated feedback on lesson efficiency. We analyzed the time it took tutors to complete lessons among learners in the TOT condition (those receiving LLM feedback) compared to the Control and ITT conditions, who did not receive LLM feedback. The analysis focused on log-transformed time data to normalize the distribution and ensure desirable outcome distributions for regression modeling. A linear mixed model was fitted with log-transformed time taken as the dependent variable, AI-generated feedback as the primary predictor, and random intercepts for learners and lessons to account for nested data, similar to RQ1.

Results indicate no statistically significant effect of receiving LLM feedback on the time taken to complete lessons (Estimate = 0.00, 95\% CI [-0.08, 0.08], p = .999). Indeed, the conditions were surprisingly similar, with learners receiving feedback taking an average of 5.75 minutes to complete lessons, while others took an average of 5.59 minutes, which is a difference of merely 9 seconds. The intraclass correlation coefficient (ICC) shows that 33\% of the variance in time taken was attributable to differences between learners, while only 2\% of the variance was due to differences between lessons. These findings suggest that receiving LLM feedback did not meaningfully alter the time spent on lessons, and individual differences were the primary driver of variation in lesson efficiency.

\subsection{RQ3: Learner Perceptions of LLM-Generated Feedback}

We analyzed learner perceptions of LLM feedback based on an optional, single-survey response collected at the tutor level. Among the 568 lessons completed (TOT), 332 learners chose to provide feedback on their perceptions of the feedback. The results show that the vast majority of learners classified it as helpful 312 (93.98\%), while 14 (4.22\%) found it not helpful, and only  6 (1.81\%) categorized it as incorrect. These findings showed overwhelmingly positive perceptions.

\section{Discussion}

\textbf{Small learning gains from LLM-generated feedback, moderated by learners' tendency to seek support.} Learners with a higher propensity to request LLM-generated feedback performed significantly better on posttest compared to those who did not. After adjusting for this selection effect, the main effect of LLM-generated feedback was positive, though there was heterogeneity across lessons. Given the significant positive effect among the group that used the feedback, we conclude that learners most likely to seek feedback learned more, which is likely associated with receiving LLM-generated feedback. In all cases, we estimate the true effect of LLM feedback to be small and possibly dependent on the specific content being practiced, warranting further research. These findings are important as they show improvements over a realistic existing system that was already providing feedback without LLMs (see Fig. 1).

Besides the question of whether LLM-generated feedback is effective or not, the key lesson learned from the present study is that even if the LLM feedback was effective, its effectiveness hinges on the willingness of learners to request and engage with it. Learners who actively engage with the feedback derive greater benefits, yet this engagement may also reflect preexisting learning differences or greater intrinsic motivation. In line with prior research on scaffolding in tutoring systems \cite{aleven2006toward}, it is possible that proactive help-seeking behaviors may constrain the broader impact of LLM-generated feedback. Research on help-seeking behavior has shown that learners who ask for help more effectively, particularly by recognizing when and how to seek help, experience significant improvements in learning outcomes \cite{aleven2006toward,roll2011improving}. It is possible, though subject to future research, that, if learners were required to request LLM feedback benefits would be larger. %

\textbf{LLM-generated feedback does not take more time.} Learners engaging with LLM feedback did not take statistically significantly more time compared to the control condition of not receiving such feedback. This suggests that providing LLM-generated feedback to learners is a low-cost instructional addition to short scenario-based lessons or even professional development training for teachers, tutors, and alike. However, in some cases, the feedback generated was rather long, making it more likely that not all learners are reading the feedback. Nevertheless, upon adjusting for self-selection, there were learning benefits to receiving LLM feedback, so at least some learners process the feedback. We suspect that the lack of completion time differences could be due to increased fluency through marginal learning: Learners benefit from receiving feedback and completing the posttest in a short period of time due to increased fluency. 

Another alternative explanation for the lack of completion time differences by condition is that learners receiving LLM feedback (\textit{treatment-on-the-treated} condition) used generative AI to generate their open responses at posttest, shortening the time required to produce a response and producing marginally higher posttest performance. A follow-up analysis revealed that learners who received LLM feedback during instruction (in Q1, see Fig. 1) produced significantly longer responses at posttest (in Q9), six words longer on average. While longer responses may indicate better learning, it is also possible that some learners copied or adapted the LLM-generated feedback from Q1, which could have inflated their performance. According to Fan et al. \cite{fan2024beware}, the use of LLMs may encourage learner dependence on the technology contributing to metacognitive “laziness,” hindering learners’ ability to generate, critically reflect on, and effectively learn from their responses when needed. Although we cannot rule out such misuse of genAI, we posit that such cases were rare, as our lessons are low-stakes assessments and take learners a short time to complete (averaging <6 mins).

\textbf{Learners found the LLM-generated feedback helpful.} Most of the learners in the LLM feedback condition (\textit{treatment-on-the-treated}) rated the feedback as \textit{helpful}, echoing the findings of Kinder et al. \cite{kinder2025effects}. 
While this positive reception is encouraging, we also acknowledge the potential for self-selection bias: learners who chose to engage with the feedback may already have been more motivated or open to support. However, the overwhelmingly positive responses warrant another interpretation: Learners’ positive ratings may reflect both genuine appreciation of the feedback and a possible influence of social desirability. The lack of optional textual comments leaves open whether learners had nuanced reactions they did not express. Nonetheless, the overwhelmingly positive ratings point to a strong initial acceptance of LLM-generated feedback as a useful instructional tool. While liking does not always equate to learning \cite{sung2012graphics}, the positive perceptions observed here suggest that LLM feedback has the potential to enhance learning—particularly when learners are inclined to engage with it.

\section{Limitations, Future Work, and Conclusion}

There are several limitations. We strived to correct for possible selection bias within the \textit{treatment-on-the-treated} group but there is a possibility that bias still exists, given that the propensity scoring method in itself is imperfect. Calls to the OpenAI API failed occasionally, with 15.7\% of requests for LLM feedback failing, with some tutors assigned the LLM-generated feedback condition opting for feedback but not able to receive it. We suspect that this reduced the overall use of LLM feedback, but it may also have undermined learner trust in the feedback, which could be investigated more closely in future research. Future work will analyze log data for time spent reading LLM feedback versus skipping ahead and examine carryover effects in AI-generated responses. Future work will also explore why lessons on motivation showed stronger effects, while equity lessons did not—potentially due to content complexity or feedback alignment.

In sum, this study highlights the promise of LLMs for generating explanatory feedback to enhance learning outcomes in scenario-based training. As our key empirical finding, the effectiveness of such feedback depends not only on its quality, but also on learners’ willingness to seek and engage with it. Analyzing over 2,000 lesson completions, we found that learners who opted to receive LLM-generated feedback consistently outperformed those who did not. After adjusting for self-selection using propensity scoring methods, we observed a small effect of LLM feedback on posttest performance. Notably, this learning benefit came without a significant increase in lesson completion time. We contribute an open dataset, annotation rubrics, and LLM scoring prompts, enhancing future research, which may explore how to maximize the impact of LLM feedback through better and more adaptive help-seeking support.

\section*{Acknowledgments}
This work was made possible with the support of the Learning Engineering Virtual Institute. The opinions, findings and conclusions expressed in this material are those of the authors.

\bibliographystyle{splncs04}
\bibliography{main} %

\end{document}